\documentclass[letterpaper, 10 pt, conference]{ieeeconf}  % Comment this line out if you need a4paper

\IEEEoverridecommandlockouts                              % This command is only needed if 
\usepackage{amsmath} % assumes amsmath package installed
\usepackage{amssymb}  % assumes amsmath package installed
\usepackage{graphicx} % for including graphics
\usepackage{booktabs}
\usepackage{cuted}
\usepackage{float}
\usepackage{capt-of} % 用于在非浮动环境中生成图表标题
\usepackage{mathtools}
\usepackage{tabularx}
\usepackage{xcolor}
\usepackage{dsfont}
\usepackage{url}
\usepackage[hidelinks,breaklinks=true]{hyperref}
\title{\LARGE \bf
REACH: An \underline{R}L-based Hi\underline{e}r\underline{a}r\underline{ch}ical Humanoid Crawling Framework for Adaptable Confined Space Traversal
}

\author{Shengzhang Xu$^{1}$, Yuning Guo$^{1}$, Weiyi Hu$^{1}$,
Yifeng Xu$^{1}$, Zhongyu Li$^{2}$, and He Kong$^{1,*}$}

\begin{document}

\twocolumn[{%
\renewcommand\twocolumn[1][]{#1}%
\maketitle
\begin{center}
\includegraphics[width=0.9\textwidth]{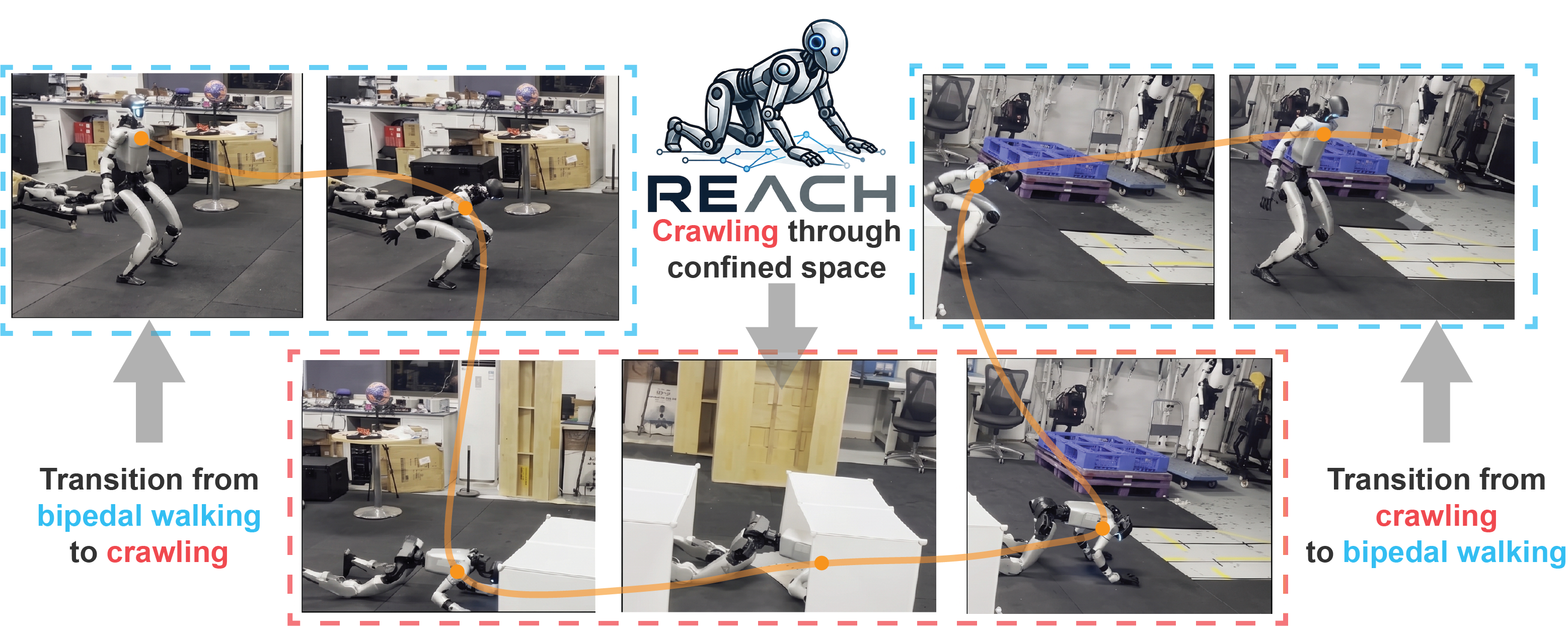}
\captionof{figure}{We introduce REACH, a modular, adaptable, RL-based hierarchical framework for humanoid crawling, enabling the Unitree G1 to transition smoothly between bipedal walking and crawling while traversing confined spaces.}
\label{fig:tran_refine_v1}
\end{center}%
}]

% Footnotes must be inserted OUTSIDE the boxed optional argument above.
% One unnumbered note preserves the ieeeconf footnote styling and places
% the affiliations at the bottom of the first (left-hand) column.
\begingroup
\renewcommand{\thefootnote}{}%
\footnotetext{%
$^{1}$S. Xu, Y. Guo, W. Hu, Y. Xu, and H. Kong are with the  Southern University of Science and Technology, Shenzhen 518055, China. Emails: jasperhsu928@hotmail.com, [12311718; huwy2023]@mail.sustech.edu.cn, 
% huwy2023@mail.sustech.edu.cn, 
yifengxu@umich.edu, kongh@sustech.edu.cn. \par
$^{2}$Z. Li is with the Chinese University of Hong Kong, Hong Kong, China, zhongyuli@cuhk.edu.hk.\par
$^{*}$Corresponding author: H. Kong. This work was supported
in part by the National Natural Science Foundation of China under Grant 62550003, in part by the Science, Technology, and Innovation Commission of Shenzhen Municipality, China under Grant JCYJ20240813094212017, in part by the Shenzhen Science and Technology Program under Grant KQTD20221101093557010, in part by Guangdong Science and Technology Program under Grant 2024B1212010002, in part by InnoHK initiative of the Innovation and Technology Commission of the Hong Kong Special Administrative Region Government via the Hong Kong Centre for Logistics Robotics.
}
\endgroup

\begin{abstract}
Humanoid locomotion in confined spaces requires continuous height adaptation across multiple locomotion modes, yet most existing approaches remain limited to height-adjustable bipedal walking. We present REACH, an RL-based hierarchical framework that enables a Unitree G1 humanoid to traverse height-constrained environments through variable-height bipedal walking and crawling. The learned policies regulate body height over complementary command ranges of $1.27$--$0.50$~m in bipedal locomotion and $0.50$--$0.20$~m in crawling, extending traversal beyond the clearance limits of crouched walking. To address contact-rich crawling dynamics, we employ a history-conditioned latent representation that captures terrain and interaction context from proprioceptive observations. A bidirectional transition policy further connects crouched bipedal locomotion and crawling across large kinematic and contact discontinuities. These skills are integrated with a LiDAR-based planner that converts point clouds into collision-free three-dimensional paths and locomotion commands, including proactive height targets. Real-world experiments demonstrate accurate height and velocity tracking, robust terrain traversal, reliable mode transitions and confined-space traversal with the proposed pipeline. Code and videos are available at \href{https://aislab-sustech.github.io/REACH/}{\textcolor{blue}{\underline{https://aislab-sustech.github.io/REACH/}}}.
\end{abstract}

%%%%%%%%%%%%%%%%%%%%%%%%%%%%%%%%%%%%%%%%%%%%%%%%%%%%%%%%%%%%%%%%%%%%%%%%%%%%%%%%
\section{INTRODUCTION}
Confined-space traversal is  essential to extending humanoid robots from structured settings to environments with 
% overhanging structures and 
narrow passages \cite{gu2026humanoid,miki2024}. However, the existing literature remains dominated by bipedal gaits \cite{huang2018}, with torso height either fixed or adjusted only for crouched walking. This restriction prevents existing systems from accessing severely constrained spaces. In recent years, reinforcement learning (RL) has substantially enhanced humanoid mobility and robustness \cite{li2025reinforcement, li2021robust, ren2026stubborn} or posture adaptation under complex terrain and overhead constraints \cite{radosavovic2024learning,song2025gait,xue2026collision,ben2026gallant}. 
% Perception-aware methods further enable posture adaptation under overhead constraints \cite{xue2026collision,ben2026gallant}. 
Nevertheless, most approaches remain bipedal and are limited by leg workspace: 
aggressive crouching reduces maneuverability and balance margins. 

However, extending a conventional humanoid from bipedal walking to another motion modality, e.g., crawling, introduces a few challenges. First, the two modes differ substantially in configuration--directly switching between two independently trained policies requires the robot to pass through dynamically sensitive intermediate states and can produce discontinuous actions, imbalance, or falls. Second, crawling is intrinsically contact-rich:  body segments such as the palms, wrists, knees, feet interact with the environment, generating impacts and disturbances that are difficult to model, leading to severe sim-to-real gaps. 

% discrepancies are particularly detrimental during multi-contact locomotion. 

Towards a practical solution for large-range height adaptation, robust multi-contact control, smooth transitions between locomotion modes,
% , and a modular interface through which perception and planning methods can command the learned skills. Towards this goal, 
we present REACH (see Fig. \ref{fig:tran_refine_v1}), an RL-based hierarchical framework for adaptable confined-space traversal using an unmodified Unitree G1 humanoid. Inspired by latent terrain-inference approaches such as DreamWaQ \cite{nahrendra2023dreamwaq}, we design a history-conditioned latent adaptation module tailored to contact-rich humanoid crawling. The module extracts terrain and interaction context from proprioceptive histories, improving robustness to multi-contact disturbances and real-world deployment conditions. A bidirectional motion-tracking policy connects crouched walking and crawling, enabling stable transitions across large kinematic and contact discontinuities. At the upper level, an adapted EGO-Planner~\cite{c21} converts onboard LiDAR observations into collision-free 3D paths and proactive height commands. This hierarchical decomposition separates perception and planning from locomotion, allowing integration with other navigation methods. Our main contributions are:
\begin{itemize}
\item An adaptable locomotion framework that extends locomotion beyond the clearance limits of crouched walking. More specifically, the proposed framework enables variable-height bipedal walking (from 1.27 to 0.50~m) and crawling with height commands from 0.50 to 0.20~m (in experiments, traversal of clearances as low as $0.40$~m).
\item A history-conditioned latent adaptation mechanism and a bidirectional
  transition policy for contact-rich crawling and bipedal--crawling switching.
\item A practical planner--controller hierarchy that interfaces learned
  skills with LiDAR-based planning, validated successful real-world confined-space experiments.
% \item The proposed framework enables
%   variable-height bipedal walking (from 1.27 to 0.50~m) and crawling with height commands
%   from 0.50 to 0.20~m (in experiments, traversal of clearances as low as $0.40$~m).
\end{itemize}

\begin{figure*}[t]
    \centering
    \includegraphics[width=0.91\textwidth]{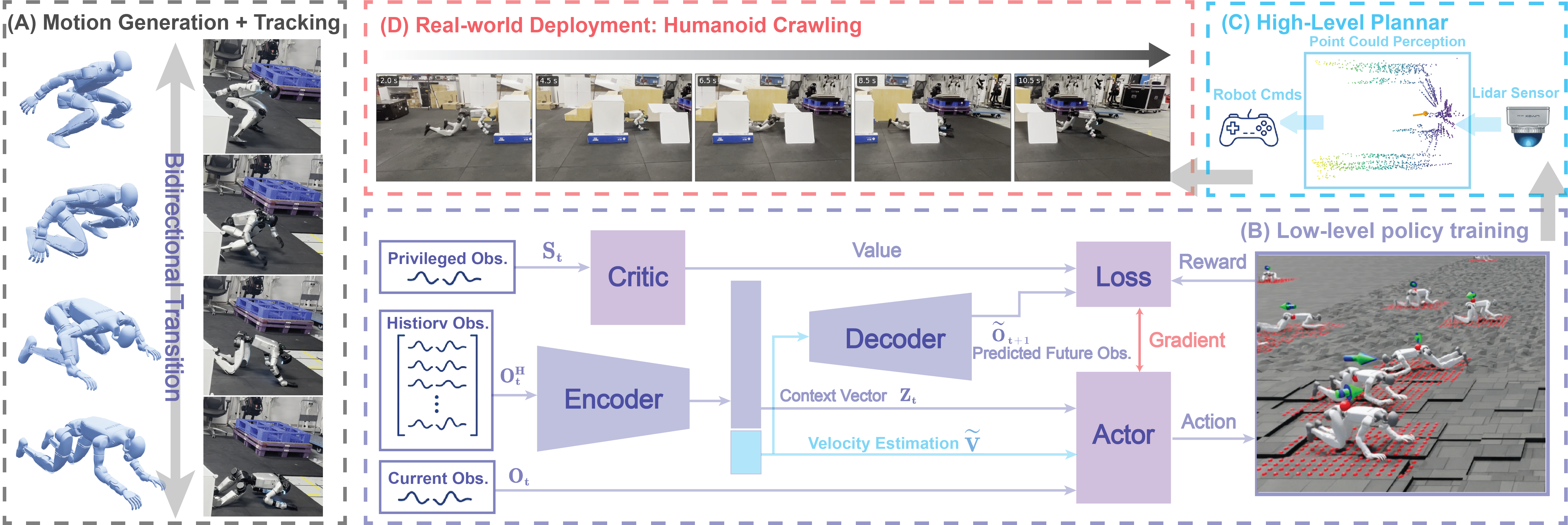}
    \caption{Overview of REACH. (A) A bidirectional motion-tracking policy connects crouched bipedal locomotion and crawling. (B) The history-conditioned asymmetric actor--critic encodes proprioceptive histories, predicts future observations and body velocity, and uses privileged information only for critic training. (C) A model-based planner converts LiDAR point clouds into locomotion and height commands. (D) Real-world trials demonstrate crawling through consecutive low-clearance obstacles.}
    \label{fig:main_v1}
\end{figure*}

\section{Related Work}
\label{sec:related_work}

\subsection{Humanoid Locomotion under Complex Terrains}

Recent learning-based humanoid locomotion methods increasingly incorporate exteroceptive perception to anticipate geometric constraints, using depth images \cite{zhu2026hiking,zhang2026rpl}, LiDAR-derived elevation maps \cite{long2025learning,he2025attention,zhang2026ame2}, or RGB observations \cite{liu2026geoloco}. These approaches improve mobility over rough, deformable, and discontinuous terrain, while model-informed objectives further enhance stability. Xie et al. \cite{xie2025humanoid} introduce a Zero-Moment Point-based reward for balance recovery, whereas Dai et al. \cite{dai2026walk} incorporate Control Lyapunov Function constraints for traversing sparse footholds. Attention-based terrain encoders additionally support accurate foothold selection \cite{he2025attention,zhang2026ame2}. Related systems achieve robust blind locomotion through large-scale reinforcement learning \cite{radosavovic2024learning} and terrain-aware traversal of stairs and gaps by combining online reconstruction with adaptive gait control \cite{song2025gait}. However, these methods primarily assume vertically unconstrained environments and remain centered on bipedal locomotion, limiting their applicability to severely narrow passages.

\subsection{Humanoid Postural Adaptation for Confined Spaces}

Confined-space traversal requires substantial posture adaptation under low ceilings and overhanging obstacles. Perception-driven methods adapt humanoid posture online: HumanoidPF \cite{xue2026collision} enables crouching and obstacle negotiation, while Gallant \cite{ben2026gallant} uses voxel representations and reinforcement learning under overhead constraints. Both remain predominantly bipedal, however, and aggressive crouching is limited by leg workspace, reduced maneuverability, and degraded disturbance rejection. Specialized low-profile platforms and hardware adaptations further ease confined locomotion \cite{schperberg2026,huang2018}, but their morphological assumptions do not transfer directly to human-sized humanoids. Bridging large-range height adaptation with contact-rich crawling on a standard humanoid therefore remains an open challenge.

\subsection{Motion Tracking and Multi-Contact Crawling}

Motion tracking and imitation learning provide a complementary route to whole-body humanoid skills. BeyondMimic \cite{liao2025beyondmimic} demonstrates versatile control by tracking richly generated motions, while generative models such as Kimodo \cite{Kimodo2026} synthesize controllable whole-body references. We adopt this tracking paradigm only for bipedal--crawling mode switching: a single bidirectional tracker follows generated crawl-to-squat references (and their reverse), rather than building a general skill library. Separately, low-profile crawling has long been studied with model-based gait design \cite{brooks2015}, compliant or specialized hardware \cite{huang2018,ma2026mash}, and morphologies that favor multi-contact support \cite{schperberg2026}. In contrast, our method learns command-conditioned crawling on an unmodified Unitree G1, uses history-conditioned latent adaptation for contact-rich disturbances, and exposes a modular interface to LiDAR-based planning.

\section{METHODOLOGY}

\iffalse
\subsection{Problem Statement}

Humanoid confined-space traversal requires continuous height regulation beyond crouched bipedal walking, together with contact-rich quadrupedal crawling and reliable switching between the two modes. We address this problem with a modular, command-conditioned hierarchy: a shared reinforcement learning architecture for variable-height walking and crawling (Sec.~\ref{sec:shared_architecture}), a bidirectional transition policy that connects their boundary postures (Sec.~\ref{sec:trans}), and a LiDAR-based planner that issues velocity and height commands to the learned skills (Sec.~\ref{sec:perception_planning}).
\fi

\subsection{RL-based Variable-Height Locomotion}
\label{sec:shared_architecture}

As shown in Fig. \ref{fig:main_v1}, we employ a shared architecture to train separate variable-height bipedal and crawling policies. The architecture combines an asymmetric actor--critic formulation with a history-conditioned context estimator, adopted from \cite{nahrendra2023dreamwaq}. Both policies share the same deployable observation and network interface, enabling command execution across locomotion modes without privileged sensing on the physical robot. The actor maps the current proprioceptive observation, estimated base velocity, and latent context to a 29-dimensional joint action $\mathbf{a}_t \sim
\pi_{\psi}
\left(
\mathbf{a}_t \mid
\mathbf{o}_t,\tilde{\mathbf{v}}_t,\mathbf{z}_t
\right),$
% \begin{equation}
% \mathbf{a}_t \sim
% \pi_{\phi}
% \left(
% \mathbf{a}_t \mid
% \mathbf{o}_t,\tilde{\mathbf{v}}_t,\mathbf{z}_t
% \right),
% \end{equation}
where the 99-dimensional observation is $\mathbf{o}_t =
[\boldsymbol{\omega}_t,\mathbf{g}_t,\mathbf{c}_t,\boldsymbol{\Phi}_t,
\boldsymbol{\theta}_t,\dot{\boldsymbol{\theta}}_t,
\mathbf{a}_{t-1}]^T.$
% \begin{equation}
% \mathbf{o}_t =
% [\boldsymbol{\omega}_t,\mathbf{g}_t,\mathbf{c}_t,
% \boldsymbol{\theta}_t,\dot{\boldsymbol{\theta}}_t,
% \mathbf{a}_{t-1}]^T.
% \end{equation}
Here, $\mathbf{c}_t=[v_x^{\mathrm{cmd}},v_y^{\mathrm{cmd}},
\omega_z^{\mathrm{cmd}},h^{\mathrm{cmd}}]$ specifies planar velocity, yaw rate, and target height, and $\boldsymbol{\Phi}_t=[\sin(2\pi\phi_t^{\mathrm{LF}}),\cos(2\pi\phi_t^{\mathrm{LF}})]$ encodes the gait clock of Eq.~\eqref{eq:walk_gait_phase}; it is set to zero for the phase-free crawling policy. A context estimator processes the five-step history
$\mathbf{o}_t^H=[\mathbf{o}_t,\ldots,\mathbf{o}_{t-4}]^T$.
Its encoder produces the latent context $\mathbf{z}_t$ and estimated base velocity $\tilde{\mathbf{v}}_t$, while a decoder reconstructs the subsequent observation. Joint velocity-estimation and future-observation reconstruction objectives encourage $\mathbf{z}_t$ to encode terrain properties, contact events and dynamics that cannot be recovered from a single observation.

% \begin{figure*}[t]
%     \centering
%     \includegraphics[width=0.95\textwidth]{images/HiQ_final_main_seq.png}
%     \caption{To be adjusted}
%     \label{fig:main_v1}
% \end{figure*}

\subsubsection{Bipedal Walking}

For training, the critic receives $\mathbf{s}_t^{\mathrm{walk}} =
[\mathbf{o}_t,\mathbf{v}_t,\mathbf{h}_t^{\mathrm{scan}},
\mathbf{p}_t^{\mathrm{feet}},\mathbf{v}_t^{\mathrm{feet}},
\mathbf{f}_t^{\mathrm{feet}},h_t^{\mathrm{root}}]^T,$
% \begin{equation}
% \mathbf{s}_t^{\mathrm{walk}} =
% [\mathbf{o}_t,\mathbf{v}_t,\mathbf{h}_t^{\mathrm{scan}},
% \mathbf{p}_t^{\mathrm{feet}},\mathbf{v}_t^{\mathrm{feet}},
% \mathbf{f}_t^{\mathrm{feet}},h_t^{\mathrm{root}}]^T,
% \end{equation}
which augments proprioception with terrain and foot-contact information available only in simulation.

\subsubsection{Crawling}

The crawling policy retains the same actor and context estimator, while the critic explicitly represents the four hand-and-knee contacts $\mathbf{s}_t^{\mathrm{crawl}} =
[\mathbf{o}_t,\mathbf{v}_t,\mathbf{h}_t^{\mathrm{scan}},
\mathbf{p}_t^{\mathrm{h\&k}},\mathbf{v}_t^{\mathrm{h\&k}},
\mathbf{f}_t^{\mathrm{h\&k}},h_t^{\mathrm{root}}]^T.$
% \begin{equation}
% \mathbf{s}_t^{\mathrm{crawl}} =
% [\mathbf{o}_t,\mathbf{v}_t,\mathbf{h}_t^{\mathrm{scan}},
% \mathbf{p}_t^{\mathrm{h\&k}},\mathbf{v}_t^{\mathrm{h\&k}},
% \mathbf{f}_t^{\mathrm{h\&k}},h_t^{\mathrm{root}}]^T.
% \end{equation}
This 308-dimensional input exposes the critic to contact-rich crawling dynamics, while the latent estimator helps the actor compensate for impacts, uncertain contacts, and modeling errors using onboard proprioception alone. Shared network interfaces therefore provide a common learning structure, whereas mode-specific privileged observations and rewards specialize training of the two locomotion policies.
\subsubsection{Reward Design}
Although the walking and crawling policies share the same command-conditioned learning architecture, they solve different contact-control problems. For mode $m\in\{\mathrm{walk},\mathrm{crawl}\}$,
\begin{equation}
r_t^{m}
=r_{\mathrm{cmd},t}^{m}
+r_{h,t}^{m}
+r_{\mathrm{post},t}^{m}
+r_{\mathrm{gait},t}^{m}
+r_{\mathrm{reg},t}^{m},
\label{eq:reward_decomposition}
\end{equation}
command tracking and regularization are shared, while height reference, posture prior, and gait shaping differ (see Appendix Tables~\ref{tab:bipedal_rewards_appendix} and~\ref{tab:crawl_rewards_appendix} for full details of reward design). We highlight the two distinctions that most affect confined-space behavior. Let $\bar{h}^{\mathrm{scan}}=\operatorname{mean}(\mathbf{h}^{\mathrm{scan}})$ and $\mathbf{c}_{\mathrm{vel}}^{\mathrm{cmd}}=[v_x^{\mathrm{cmd}},v_y^{\mathrm{cmd}},\omega_z^{\mathrm{cmd}}]^\top$.

\paragraph{Mode-specific height regulation.}
Both policies interpret $h^{\mathrm{cmd}}$ relative to local terrain, but track different body points:
\begin{align}
r_h^{\mathrm{walk}}
&=\exp\!\left[-25
\left(h^{\mathrm{head}}-\bar{h}^{\mathrm{scan}}-h^{\mathrm{cmd}}\right)^2\right],
\label{eq:walk_height_reward}\\
r_h^{\mathrm{crawl}}
&=\exp\!\left[-\frac{
\left(h^{\mathrm{base}}-\bar{h}^{\mathrm{scan}}-h^{\mathrm{cmd}}\right)^2}{0.15^2}\right].
\label{eq:crawl_height_reward}
\end{align}
Walking uses head height as an overhead-clearance proxy, with torso-pitch constraints (see Table~\ref{tab:bipedal_rewards_appendix}) preventing trivial satisfaction by upper-body tilt. Crawling uses base height as a more stable clearance measure when the torso is nearly horizontal, and the head moves with arm placement.

\paragraph{Phase-indexed walking versus phase-free crawling.}
Bipedal walking uses an explicit periodic contact schedule. For each foot $l\in\mathcal{L}_{\mathrm{feet}}$,
\begin{equation}
\begin{aligned}
\phi_t^l
&=\left(\frac{t\bmod T}{T}+o^l\right)\bmod1,\\
T&=0.8~\mathrm{s},\qquad
(o^{\mathrm{LF}},o^{\mathrm{RF}})=(0,0.5),
\end{aligned}
\label{eq:walk_gait_phase}
\end{equation}
with desired stance $D_t^l=\mathds{1}(\phi_t^l<0.55)$ and measured contact $C_t^l=\mathds{1}(F_{z,t}^l>1)$. Matching these states yields
\begin{equation}
\begin{aligned}
r_{\mathrm{gait}}^{\mathrm{walk}}
&=\mathds{1}\!\left(\|\mathbf{c}_{\mathrm{vel}}^{\mathrm{cmd}}\|>0.1\right)
\frac{1}{2}\sum_{l\in\mathcal{L}_{\mathrm{feet}}}
\mathds{1}(C_t^l=D_t^l),
\end{aligned}
\label{eq:walk_gait_reward}
\end{equation}
encouraging alternating support with a duty factor close to normal walking~\cite{winter1984kinematic}. A fixed four-limb phase clock is less suitable for crawling, where timing varies with height, terrain, and hand--knee impacts. Instead, for $\mathcal{L}_{\mathrm{h\&k}}=\{\mathrm{LH},\mathrm{RH},\mathrm{LK},\mathrm{RK}\}$ and $C_t^l=\mathds{1}(F_{z,t}^l\geq20)$, we penalize unsafe support combinations
\begin{align}
p_{\mathrm{pair}}
&=\mathds{1}(C^{\mathrm{LH}}+C^{\mathrm{RH}}=0)
+\mathds{1}(C^{\mathrm{LK}}+C^{\mathrm{RK}}=0),
\label{eq:crawl_pair_penalty}\\
p_{\mathrm{ipsi}}
&=\mathds{1}(C^{\mathrm{LH}}+C^{\mathrm{LK}}=0)
+\mathds{1}(C^{\mathrm{RH}}+C^{\mathrm{RK}}=0),
\label{eq:crawl_ipsi_penalty}
\end{align}
which preserve longitudinal and lateral support and favor diagonal hand--knee coordination without prescribing an exact stance sequence. Mode-duration and left--right timing terms further suppress contact chatter and asymmetric cycles:
\begin{equation}
\begin{aligned}
r_{\mathrm{mode}}
&=\mathds{1}\!\left(\|\mathbf{c}_{\mathrm{vel}}^{\mathrm{cmd}}\|>0.1\right)
\min\!\left(
\min_{l\in\mathcal{L}_{\mathrm{h\&k}}}\Delta_t^l,0.4
\right),
\end{aligned}
\label{eq:crawl_mode_time_reward}
\end{equation}
\begin{equation}
p_{\mathrm{sym}}
=
\frac{\mathds{1}\!\left(
\|\mathbf{c}_{\mathrm{vel}}^{\mathrm{cmd}}\|>0.1
\right)}{4}
\sum_{q\in\{\mathrm{H},\mathrm{K}\}}
\sum_{\tau\in\{\mathrm{lc},\mathrm{la}\}}
\left|t_{\tau}^{\mathrm{L}q}-t_{\tau}^{\mathrm{R}q}\right|.
\label{eq:crawl_symmetry_penalty}
\end{equation}
where $\Delta_t^l$ is the elapsed duration of the active contact or aerial mode of limb $l$, and $t_{\mathrm{lc}}^l$, $t_{\mathrm{la}}^l$ are the most recently completed contact and aerial durations. Together these terms define a \emph{phase-free} crawling objective---stable support and balanced timing without a global gait clock---in contrast to the phase-indexed walking reward.

\subsubsection{Training Settings}

The two modes share the planar-velocity range but use different yaw-rate and height ranges:
\begin{align*}
\mathcal{C}_{\mathrm{walk}}:
&\quad v_x^{\mathrm{cmd}}\in[-0.6,0.8],\quad
v_y^{\mathrm{cmd}}\in[-0.6,0.6],\nonumber\\
&\quad \omega_z^{\mathrm{cmd}}\in[-1.0,1.0],\quad
h^{\mathrm{cmd}}\in[0.50,1.27],\\[1mm]
% \label{eq:walk_cmd_range}\\[1mm]
\mathcal{C}_{\mathrm{crawl}}:
&\quad v_x^{\mathrm{cmd}}\in[-0.6,0.8],\quad
v_y^{\mathrm{cmd}}\in[-0.6,0.6],\nonumber\\
&\quad \omega_z^{\mathrm{cmd}}\in[-0.75,0.75],\quad
h^{\mathrm{cmd}}\in[0.20,0.50].
% \label{eq:crawl_cmd_range}
\end{align*}
Linear velocity, yaw rate, and height are expressed in $\mathrm{m\,s^{-1}}$, $\mathrm{rad\,s^{-1}}$, and meters, respectively. The height commands $h^{\mathrm{cmd}}\in[0.50,1.27]$~m and $h^{\mathrm{cmd}}\in[0.20,0.50]$~m are used only during policy training. They are terrain-relative targets for head height in walking and base height in crawling, and do not represent the robot's physical height. For both modes, velocity commands are resampled every $5$--$10$~s and height commands every $2$--$10$~s. Both policies use the same terrain curriculum over box arrays, random rough ground, undulating terrain, slopes, and platforms, with difficulty increasing gradually.
% to the successfully traversed distance. 
The termination criteria remain mode-specific. Walking terminates upon torso--terrain contact ($F_z^{\mathrm{torso}}>1$), whereas crawling terminates when the torso crosses the rollover criterion ($g_x^{\mathrm{torso}}<0$). Combined with the termination penalty, these criteria encourage fall avoidance in walking and stable four-contact support in crawling.

\subsection{Bidirectional Bipedal--Crawling Transition}
\label{sec:trans}
The bipedal and crawling policies are trained independently and therefore operate around different posture and contact distributions. Direct switching is unreliable because neither policy is trained to recover from the intermediate states between a deep squat and four-limb support. We connect the two skills using a generated reference motion and a single bidirectional tracking policy, as illustrated in Fig.~2(A). This decomposition makes the framework extensible: a new locomotion skill can be connected by generating a transition between compatible endpoint postures and training a tracker, without modifying the existing low-level policies.

\subsubsection{Reference Motion Generation}

We use Kimodo~\cite{Kimodo2026} to generate a coherent transition from quadrupedal crawling to a deep squat. Compared with collecting and retargeting motion-capture data, motion generation provides direct control over the terminal postures, allowing them to match the boundary states of the learned locomotion policies. In our implementation, the crawling endpoint corresponds to the stationary crawling policy at its maximum commanded height of $0.5$~m, whereas the squat endpoint matches the stationary bipedal policy at its minimum height of $0.5$~m.

The generated motion lasts $5$~s at $30$~Hz. Whole-body keyframes guide the sequence from the initial crawl, through a compact posture with the legs advanced beneath the body, to a single-foot--single-knee support state with both hands lifted, and finally to a stabilized deep squat, as shown in Fig.~2(A). The text prompt specifies the same progression: sequential leg advancement, arm retraction, and successive knee lift. These constraints preserve the intended contact order while leaving the generator free to synthesize the intervening whole-body motion. We generate the dynamically harder crawl-to-squat direction; the reverse sequence is used only as a tracking reference, while the physics-based policy learns a feasible closed-loop realization of both directions.

\subsubsection{Unified Transition Motion Tracking}

Confined-space traversal requires both crawl-to-squat and squat-to-crawl switching; because contact changes and balance constraints are direction-dependent, a one-way tracker is insufficient. We therefore train a \emph{single} transition controller, extending BeyondMimic~\cite{liao2025beyondmimic}, to realize both directions; bidirectionality therefore refers to this shared policy, not to kinematic time-reversal alone. In practice, half of the parallel environments track the fixed reference in crawl-to-squat order and the remainder track its reverse, with the playback direction resampled every $2$--$6$~s so that one policy covers both transitions and their intermediate states. We also extend the observation history from one to three steps, providing sufficient temporal context to distinguish rising toward the squat from descending toward four-limb support (Fig.~2(A)).

To smooth switching transients, the supervisor first brings the robot to the shared $0.5$~m endpoint under the active policy, then runs the closed-loop tracker along the reference, and finally transfers control to the target policy, rather than swapping policies through unsupported intermediate states. State and dynamics randomization, with permissive termination, further trains the tracker to absorb handoff mismatch and external disturbances. Under strong randomization, the standard tracking objective occasionally admits a kneeling local optimum near the squat endpoint. Strict termination can remove this behavior but substantially reduces useful training data. We instead introduce phase-gated shaping terms, summarized in Appendix Table~\ref{tab:transition_rewards_appendix}. Let $t_c$ denote the current reference-frame index. During the rising phase, head- and knee-height rewards promote completion of the deep squat, foot-level and non-foot-contact penalties stabilize the final support configuration, and hand-level shaping maintains reliable palm support before lift-off. These terms target the ambiguous contact changes without constraining the entire trajectory, preserving the diversity and robustness of the original motion-tracking objective.

\begin{figure*}[t]
    \centering
    \includegraphics[width=0.95\textwidth]{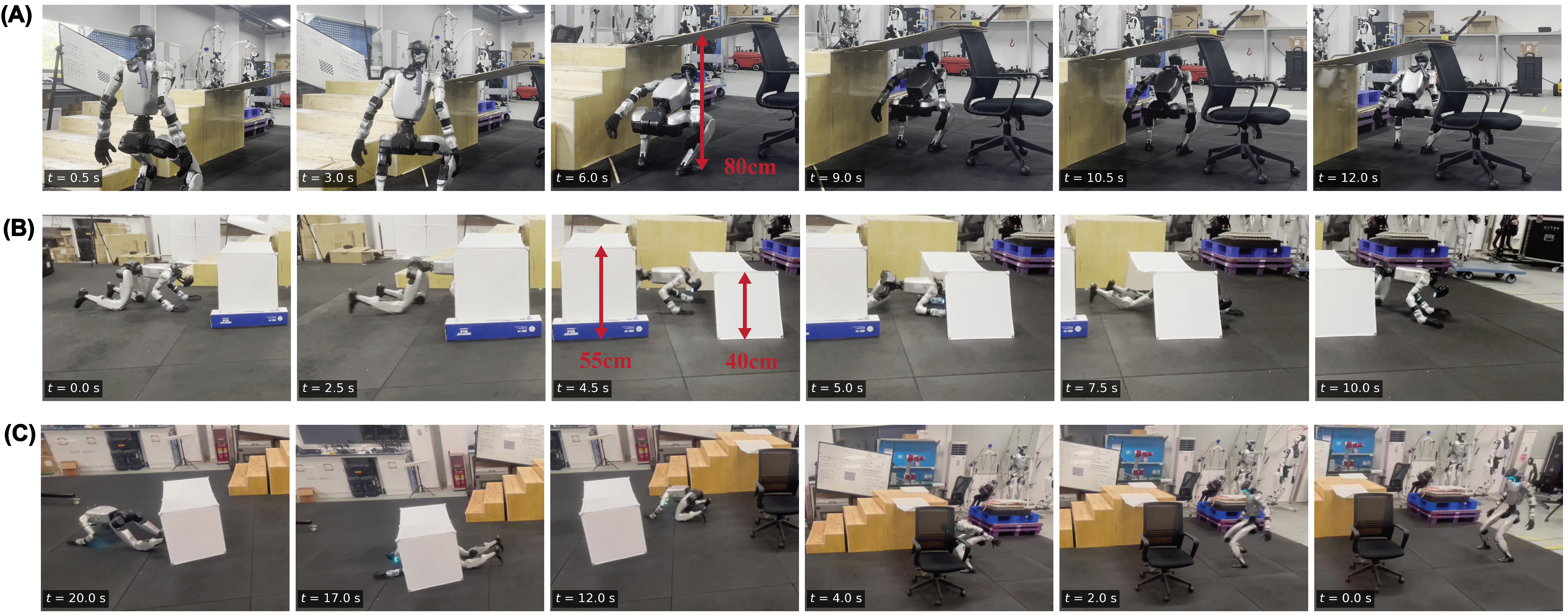}
    \caption{Real-world evaluation of the complementary locomotion capabilities enabled by REACH. (A) Variable-height bipedal walking continuously adjusts body height to traverse an 80~cm-clearance obstacle. (B) Quadrupedal crawling enables robust multi-contact traversal through passages as low as 40~cm. (C) Bidirectional transitions between bipedal standing and quadrupedal crawling enable the robot to enter and exit low-clearance obstacles under different locomotion modes.}
    \label{fig:composite}
\end{figure*}

\section{Perception and Planning}
\label{sec:perception_planning}

Figure~2(C) summarizes the hierarchical navigation framework. The model-based front end converts LiDAR measurements into a collision-free three-dimensional body path, while a lightweight tracking interface converts this path into the velocity and height commands accepted by the learned locomotion policies. A locomotion supervisor then dispatches these commands to the bipedal or crawling policy and invokes the bidirectional transition policy when the required mode changes.

% This decomposition assigns geometric reasoning to the planner and contact-rich balance control to the learned policies; consequently, none of the low-level policies must directly process point clouds or global paths.

\subsection{Spatiotemporal Point-Cloud Representation}

Each LiDAR scan is first processed by Elevation Mapping~\cite{c22} to estimate the local ground surface and separate obstacle returns from traversable terrain. A single scan, however, is incomplete because of the sensor's limited field of view and self-occlusion by the robot. We therefore transform the current obstacle points into the world frame using the pose estimated by Point-LIO~\cite{c23} and fuse them with its accumulated static map. The resulting obstacle cloud combines current measurements with previously observed geometry, retaining overhanging structures that cannot be represented by a ground-height map alone. This temporally persistent representation is supplied to the front-end A* search and the subsequent path optimizer.

\subsection{Humanoid-Constrained Spatial Planning}

We adapt EGO-Planner~\cite{c21}, originally developed for aerial robots, to produce geometric paths compatible with variable-height humanoid locomotion. Let $\mathbf{Q}=\{\mathbf{q}_i\}_{i=1}^{N_c}$ denote the path control points, where $\mathbf{q}_i=[x_i,y_i,z_i]^\top$. The original optimization $J_{\mathrm{orig}}(\mathbf{Q})
=\lambda_s J_s+\lambda_c J_c+\lambda_d J_d$ balances smoothness, collision avoidance, and dynamic feasibility.
% \begin{equation}
% J_{\mathrm{orig}}(\mathbf{Q})
% =\lambda_s J_s+\lambda_c J_c+\lambda_d J_d.
% \label{eq:ego_original_objective}
% \end{equation}
Because timing, velocity tracking, and dynamic balance are handled by the low-level policies, our planner optimizes a static spatial path rather than a dynamically feasible aerial trajectory. We therefore remove $J_d$ and introduce two humanoid-specific geometric terms: $J_{\mathrm{ours}}(\mathbf{Q})
=\lambda_s J_s+\lambda_c J_c+\lambda_h J_h
+\lambda_{ot}J_{ot}.$
% \begin{equation}
% J_{\mathrm{ours}}(\mathbf{Q})
% =\lambda_s J_s+\lambda_c J_c+\lambda_h J_h
% +\lambda_{ot}J_{ot}.
% \label{eq:humanoid_planning_objective}
% \end{equation}

\paragraph{Terrain-relative height regularization.}
The first term regulates the body height along the path while allowing vertical adaptation when it is preferable to a long horizontal detour. Let $\hat z_g(x_i,y_i)$ be the estimated ground elevation beneath $\mathbf{q}_i$, and define the planned terrain-relative height as $h_i=z_i-\hat z_g(x_i,y_i)$. We use $J_h=\sum_{i=1}^{N_c}\left(h_i-h^{\mathrm{nom}}\right)^2,$
% \begin{equation}
% J_h=\sum_{i=1}^{N_c}\left(h_i-h^{\mathrm{nom}}\right)^2,
% \label{eq:planner_height_cost}
% \end{equation}
where $h^{\mathrm{nom}}$ is the nominal walking height. In open space, this term suppresses unnecessary vertical oscillations. Near an overhead obstacle, its finite weight allows the optimizer to lower the path when the resulting height change is less costly than a lateral bypass. Importantly, expressing the cost relative to the local ground makes the planned height consistent with the height commands used to train both low-level policies.

\paragraph{Overhead traversal suppression.}
An aerial planner may pass above an obstacle whenever doing so shortens the path, even when the obstacle exceeds the humanoid's stepping capability. To suppress such fly-over solutions, we penalize path points above the obstacle footprint via $J_{ot}=\sum_{i\in\Omega_{\mathrm{top}}}
\exp\!\left(\kappa\left[z_i-z_{\mathrm{obs}}(x_i,y_i)\right]\right),$
% \begin{equation}
% J_{ot}=\sum_{i\in\Omega_{\mathrm{top}}}
% \exp\!\left(\kappa\left[z_i-z_{\mathrm{obs}}(x_i,y_i)\right]\right),
% \label{eq:planner_overhead_cost}
% \end{equation}
where $\Omega_{\mathrm{top}}$ contains control points whose horizontal coordinates lie within an obstacle footprint, $z_{\mathrm{obs}}(x_i,y_i)$ is the obstacle's top elevation, and $\kappa>0$ controls the growth rate. The cost increases rapidly above the obstacle and thus acts as a soft vertical extension of its footprint. Together with $J_c$, it directs the optimizer around solid obstacles while preserving collision-free underpasses beneath elevated structures. The two added terms therefore distinguish feasible humanoid height adaptation from the unrestricted vertical motion assumed by the original aerial planner.

\subsection{3D Path-to-Policy Interface}

The optimized path $\mathcal{P}$ is converted into low-dimensional commands by a three-dimensional Pure Pursuit tracker running at $30$~Hz. At each update, a lookahead sphere of radius $L_d$, centered at the robot, selects a forward target on $\mathcal{P}$. After transformation into the robot base frame, this target is written as ${}^{B}\mathbf{p}_L=[x_L,y_L,z_L]^\top$. The horizontal components determine the commanded planar velocity, and the local path-heading error determines the yaw-rate command; velocity-coupled lateral feedback compensates for drift during path tracking. The vertical component is converted to a terrain-relative body-height command, providing advance notice of an approaching overhead constraint because the target lies ahead of the robot rather than at its current position. The resulting interface is $\mathbf{c}_t=
\begin{bmatrix}
v_{x,t}^{\mathrm{cmd}} &
v_{y,t}^{\mathrm{cmd}} &
\omega_{z,t}^{\mathrm{cmd}} &
h_t^{\mathrm{cmd}}
\end{bmatrix}^{\!\top},$
% \begin{equation}
% \mathbf{c}_t=
% \begin{bmatrix}
% v_{x,t}^{\mathrm{cmd}} &
% v_{y,t}^{\mathrm{cmd}} &
% \omega_{z,t}^{\mathrm{cmd}} &
% h_t^{\mathrm{cmd}}
% \end{bmatrix}^{\!\top},
% \label{eq:planner_policy_command}
% \end{equation}
which exactly matches the command parameterization used during low-level policy training. The planned height also coordinates locomotion-mode selection. The bipedal and crawling policies meet at the shared switching height $h^{\mathrm{sw}}=0.5$~m: walking realizes commands above this boundary, whereas crawling realizes lower-clearance commands. When the requested height crosses the boundary, the supervisor first brings the active policy to the corresponding endpoint posture, executes the bidirectional transition policy, and then transfers control to the target locomotion policy. Commands are restricted to the training range of the active policy, preventing the planner from requesting heights or velocities outside its learned operating domain. Thus, a single geometric path jointly specifies where the robot should move, how its body height should change, and when a locomotion-mode transition is required.
% -----------------------------------------------
\section{Experiments and Discussions}
\label{sec:experiments}

We organize the evaluation around three research questions: \textbf{RQ1}, how do the low-level ablations affect curriculum progression, which is an indicator of training progress; \textbf{RQ2}, how accurately do the resulting policies jointly track velocity and height commands; and \textbf{RQ3}, which training components enable robust bidirectional transitions and sim-to-real transfer? The first two questions evaluate variable-height walking and crawling in simulation. The third separates transition-policy ablations in simulation from deployment of the full policy on hardware. Complementing these controlled studies, Fig.~\ref{fig:composite} further shows \textbf{real-world composability}: 80~cm-clearance walking, crawling as low as 40~cm, and composable skills in a single run that combines crouched walking and crawling via bidirectional transitions (Fig.~\ref{fig:composite}(C)).

\subsection{Variable-Height Locomotion}

\paragraph{Ablation design.}
For each locomotion mode, we train four policy variants in Isaac Lab~\cite{mittal2025isaaclab} using 8192 parallel environments and 4000 training iterations. All variants receive current proprioception; Table~\ref{tab:locomotion_ablations} summarizes their differences. \textsc{Full} is the proposed deployable architecture, in which the history-conditioned estimator provides the actor with an estimated body velocity $\tilde{\mathbf{v}}_t$ and latent context $\mathbf{z}_t$. \textsc{No-Idle} removes the linear- and angular-command idle penalties from \textsc{Full}. \textsc{Proprio} removes both estimated quantities and uses only current proprioception. Finally, \textsc{Priv-Scan} augments current proprioception with privileged ground-truth terrain heights $\mathbf{h}^{\mathrm{scan}}$; it is a simulation-only reference.
% because these privileged measurements are unavailable at deployment.

\begin{table}[t]
    \centering
    \caption{Ablation variants for different locomotion modes.}
    \label{tab:locomotion_ablations}
    \footnotesize
    \setlength{\tabcolsep}{4pt}
    \renewcommand{\arraystretch}{1.1}
    \begin{tabularx}{\columnwidth}{@{}Xccc@{}}
        \toprule
        \textbf{Variant}
        & \textbf{Hist. estimator}
        & \textbf{Priv. scan}
        & \textbf{Idle penalty} \\
        \midrule
        \textsc{Full}
        & $\checkmark$ & -- & $\checkmark$ \\
        \textsc{No-Idle}
        & $\checkmark$ & -- & -- \\
        \textsc{Proprio}
        & -- & -- & $\checkmark$ \\
        \textsc{Priv-Scan}
        & -- & $\checkmark$ & $\checkmark$ \\
        \bottomrule
    \end{tabularx}
\end{table}

\subsubsection{RQ1: Curriculum Progression}

Terrain difficulty increases only when the robot traverses a prescribed distance; its progression therefore measures how quickly a policy acquires command-responsive locomotion on increasingly complex terrain. Figure~\ref{fig:low_level_evaluation}(a,b) reports the curriculum level throughout training. For walking, \textsc{Full} is the first variant to exceed terrain level 4, doing so at approximately 1700 iterations. \textsc{Proprio} and \textsc{Priv-Scan} eventually reach similar levels but require substantially more training, whereas \textsc{No-Idle} remains below level 3 at the 4000-iteration budget. The crawling task provides a clearer separation: at iteration 4000, \textsc{Full} reaches approximately level 3.6 and \textsc{No-Idle} reaches approximately level 3.3 after a pronounced delay, while \textsc{Proprio} and \textsc{Priv-Scan} do not progress beyond the initial level. These comparisons isolate two effects. The gap between \textsc{Full} and \textsc{No-Idle} shows that idle penalties accelerate the acquisition of command-responsive motion. The failure of \textsc{Proprio} and \textsc{Priv-Scan} in crawling further indicates that explicit terrain heights alone do not replace the temporal context and body-velocity estimate required for contact-rich locomotion.

\subsubsection{RQ2: Joint Command Tracking}

\begin{figure}[!t]
    \centering
    \includegraphics[width=0.45\textwidth]{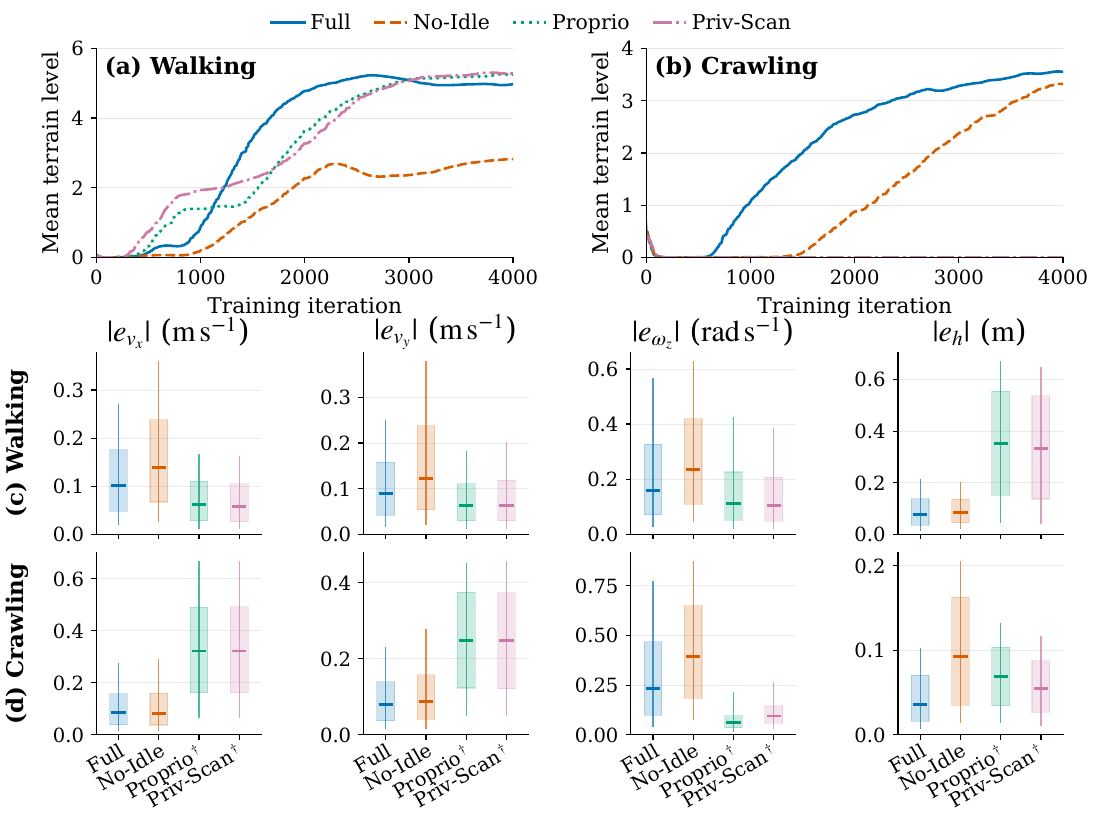}
    \caption{Low-level locomotion ablations. (a,b)~Mean terrain-curriculum level over training iterations. (c,d)~Absolute command-tracking errors (median, IQR, 10th--90th percentiles; descriptive, not confidence intervals). $\dagger$: crawling variants with no curriculum progress---low yaw-rate error should not be read as successful locomotion.}
    \label{fig:low_level_evaluation}
\end{figure}

We evaluate each trained policy in 2048 parallel environments. A command is sampled from the corresponding training range at the beginning of each rollout and held fixed. Terrain levels are set to 4 for walking and 2 for crawling. After a 1-s warm-up, we record 9~s of absolute tracking error, $e_q(t)=\left|q(t)-q^{\mathrm{cmd}}\right|,
\qquad q\in\{v_x,v_y,\omega_z,h\},$
% \begin{equation}
% e_q(t)=\left|q(t)-q^{\mathrm{cmd}}\right|,
% \qquad q\in\{v_x,v_y,\omega_z,h\},
% \label{eq:tracking_error}
% \end{equation}
where $h$ denotes the mode-specific terrain-relative height used by the corresponding low-level policy. Figure~\ref{fig:low_level_evaluation} jointly reports curriculum progression and the resulting tracking-error distributions. \textbf{In walking}, \textsc{Full} reduces all three velocity errors relative to \textsc{No-Idle}. Although \textsc{Proprio} and \textsc{Priv-Scan} exhibit smaller velocity errors, their height errors are several times larger. Rollout inspection shows that these variants often remain near an upright posture instead of realizing the commanded crouched gait; their apparently favorable velocity errors therefore do not represent successful variable-height walking. \textsc{Full} provides the best joint tracking of velocity and height. \textbf{In crawling}, \textsc{Full} has forward-velocity error comparable to \textsc{No-Idle}, while reducing lateral, yaw-rate, and height errors. \textsc{Proprio} and \textsc{Priv-Scan} show smaller yaw-rate errors, but this must be interpreted together with Fig.~\ref{fig:low_level_evaluation}(b): neither variant advances beyond the initial crawling terrain, and both have much larger translational errors. Their low yaw-rate errors therefore do not compensate for the absence of robust translational crawling. Across both evaluations, \textsc{Full} is the only deployable variant that combines rapid curriculum progression with balanced tracking of all four commands. 

\subsection{RQ3: Bidirectional Transition Robustness}

\paragraph{Ablations and protocol.}
We train three transition variants in Isaac Lab using 8192 environments and 4000 iterations. \textsc{Full (Sim.)} uses the complete motion-tracking objective, phase-gated shaping terms, dynamics randomization, and permissive termination described in the methodology. \textsc{No Shaping} removes the additional phase-gated terms and retains only the BeyondMimic objective. \textsc{Reduced Robustness} halves all randomization magnitudes except friction and halves all termination thresholds except the rollover threshold, thereby combining weaker randomization with stricter termination. The three policies are evaluated in MuJoCo~\cite{mujoco_playground} for 25 trials in each direction. Every trial begins at the shared $0.5$-m boundary posture. A squat-to-crawl trial is successful if the robot establishes stable hand--knee support without torso collision and subsequently moves under the crawling policy. Conversely, a crawl-to-squat trial must recover stable feet-only support without torso collision and subsequently walk under the bipedal policy. We additionally deploy the unchanged full policy on the physical Unitree G1 using the same initial states, criteria, and number of trials. This \textsc{Full (Real)} result is a hardware deployment of \textsc{Full (Sim.)}, not a fourth training configuration.

\begin{figure}[t]
    \centering
    \includegraphics[width=0.75\linewidth]{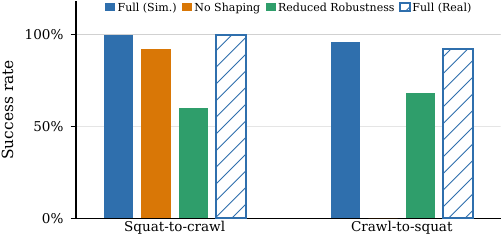}
    \caption{Bidirectional transition success over 25 trials per direction. The first three configurations are evaluated in MuJoCo; the hatched \textsc{Full (Real)} bars denote deployment of the unchanged full policy on the Unitree G1.}
    \label{fig:transition_success}
\end{figure}

\paragraph{Results.}
As shown in Fig.~\ref{fig:transition_success}, \textsc{Full (Sim.)} succeeds in $25/25$ squat-to-crawl trials and $24/25$ crawl-to-squat trials in MuJoCo, corresponding to success rates of $100\%$ and $96\%$. \textsc{No Shaping} retains $23/25$ squat-to-crawl successes but yields $0/25$ crawl-to-squat successes. This asymmetric failure supports the intended role of the shaping terms: they are specifically necessary for escaping the kneeling optimum during crawl-to-squat recovery, rather than merely improving generic tracking accuracy. \textsc{Reduced Robustness} succeeds in only $15/25$ and $17/25$ trials, showing that broad randomization and permissive termination are important for recovery from switching-state variation. On hardware, \textsc{Full (Real)} achieves $25/25$ squat-to-crawl and $23/25$ crawl-to-squat successes, or $100\%$ and $92\%$, respectively. The small decrease from simulation in the harder rising direction demonstrates that the learned bidirectional transition transfers to the physical robot while preserving reliable handover to both locomotion policies. Together with the qualitative real-world sequences in Fig.~\ref{fig:composite}, these results indicate that the full REACH stack, including variable-height walking, contact-rich crawling, and mode switching, supports robust confined-space traversal on hardware.

\section{CONCLUSIONS}

We presented REACH, an RL-based hierarchical framework that enables an unmodified Unitree G1 to traverse confined spaces through variable-height bipedal walking, crawling, bidirectional mode transitions, and a modular LiDAR planner--controller interface. Ablation studies show that history-conditioned latent adaptation improves curriculum progression and joint command tracking, particularly for contact-rich crawling. The bidirectional transition policy transfers reliably from simulation to hardware, and real-world demos demonstrate complementary walking and crawling under low clearances with planner-issued velocity and height commands.

% Do not shorten \textheight here: with a multi-page appendix and bibliography,
% \addtolength{\textheight}{-12cm} creates large blank regions and truncates
% the reference list. Reintroduce a small adjustment only on the true final page
% after the full paper is compiled, if column balancing is still needed.

\section*{APPENDIX}
\subsection*{(A)~Reinforcement Learning Reward Summary}

The signed coefficients in Tables~\ref{tab:bipedal_rewards_appendix} and~\ref{tab:crawl_rewards_appendix} are applied directly to the listed nonnegative rewards or penalties. For compactness, define $\rho_{\mathcal{J}}(\boldsymbol{\theta}^{\mathrm{ref}})=\sum_{i\in\mathcal{J}}|\theta_i-\theta_i^{\mathrm{ref}}|$ and $\beta(\theta)=\frac{1}{2}\left[
e^{-100|\min(\theta,0)|}
+e^{-100|\max(\theta-1.2,0)|}
\right].$
% \begin{equation}
% \beta(\theta)=\frac{1}{2}\left[
% e^{-100|\min(\theta,0)|}
% +e^{-100|\max(\theta-1.2,0)|}
% \right].
% \label{eq:torso_pitch_compact}
% \end{equation}
Here, $\boldsymbol{\theta}^{\mathrm{stand}}$ and $\boldsymbol{\theta}^{\mathrm{crawl}}$ are the standing and crawling reference configurations, respectively; $\mathbf{g}_{xy}$ is the horizontal component of projected gravity; and $\tau_i^{\mathrm{PD}}$ and $\tau_i$ are the PD-requested and post-saturation torques.

\begin{table}[!htbp]
\centering
\caption{Reward terms for variable-height bipedal walking.}
\label{tab:bipedal_rewards_appendix}
\footnotesize
\setlength{\tabcolsep}{2.5pt}
\renewcommand{\arraystretch}{1.12}
\begin{tabularx}{\columnwidth}{@{}p{0.28\columnwidth}Xc@{}}
\toprule
\textbf{Term} & \textbf{Definition} & \textbf{Weight} \\
\midrule
Head-height tracking & Eq.~\eqref{eq:walk_height_reward} & $2.0$ \\
Yaw-rate tracking
& $\bar{\omega}_{z,t}=\frac{1}{5}\sum_{k=0}^{4}\omega_{z,t-k}$; 
$r_{\omega}=\exp[-(\bar{\omega}_{z,t}-\omega_z^{\mathrm{cmd}})^2/0.3^2]$
& $2.0$ \\
Linear idle
& $\mathds{1}(\|\mathbf{v}_{xy}^{\mathrm{cmd}}\|>0.2\land\|\mathbf{v}_{xy}\|<0.2)$
& $-3.0$ \\
Yaw idle
& $\mathds{1}[(\omega_z^{\mathrm{cmd}}>0.2\land\omega_z<0.1)\lor(\omega_z^{\mathrm{cmd}}<-0.2\land\omega_z>-0.1)]$
& $-2.0$ \\
Arm posture
& $\rho_{\mathcal{J}_{\mathrm{arm}}}(\boldsymbol{\theta}^{\mathrm{stand}})$
& $-0.2$ \\
Waist posture
& $\rho_{\mathcal{J}_{\mathrm{waist}}}(\boldsymbol{\theta}^{\mathrm{stand}})$
& $-1.0$ \\
Hip-roll posture
& $\mathds{1}(|v_y^{\mathrm{cmd}}|\leq0.2)|\theta_{\mathrm{hip\,roll}}-\theta_{\mathrm{hip\,roll}}^{\mathrm{stand}}|$
& $-0.3$ \\
Torso-pitch range
& $\beta(\theta_{\mathrm{torso}}^{\mathrm{pitch}})$
& $1.0$ \\
High-stance torso tilt
& $\mathds{1}(h^{\mathrm{cmd}}>1.0)\|\mathbf{g}_{xy}\|^2$
& $-3.0$ \\
Phase-matched gait
& Eq.~\eqref{eq:walk_gait_reward}
& $1.0$ \\
\bottomrule
\end{tabularx}
\end{table}
\begin{table}[!htbp]
\centering
\caption{Reward terms for variable-height humanoid crawling.}
\label{tab:crawl_rewards_appendix}
\footnotesize
\setlength{\tabcolsep}{2.5pt}
\renewcommand{\arraystretch}{1.12}
\begin{tabularx}{\columnwidth}{@{}p{0.28\columnwidth}Xc@{}}
\toprule
\textbf{Term} & \textbf{Definition} & \textbf{Weight} \\
\midrule
Base-height tracking
& Eq.~\eqref{eq:crawl_height_reward}
& $3.0$ \\
Waist-pitch posture
& $|\theta_{\mathrm{waist\,pitch}}-\theta_{\mathrm{waist\,pitch}}^{\mathrm{crawl}}|$
& $-1.2$ \\
Wrist posture
& $\rho_{\mathcal{J}_{\mathrm{wrist}}}(\boldsymbol{\theta}^{\mathrm{crawl}})$
& $-0.2$ \\
Waist-roll posture
& $\mathds{1}(|v_x^{\mathrm{cmd}}|<0.3\land|\omega_z^{\mathrm{cmd}}|<0.4)|\theta_{\mathrm{waist\,roll}}-\theta_{\mathrm{waist\,roll}}^{\mathrm{crawl}}|$
& $-0.6$ \\
Hand/knee support
& Eq.~\eqref{eq:crawl_pair_penalty}
& $-0.6$ \\
Ipsilateral support
& Eq.~\eqref{eq:crawl_ipsi_penalty}
& $-0.3$ \\
Mode duration
& Eq.~\eqref{eq:crawl_mode_time_reward}
& $1.0$ \\
Timing symmetry
& Eq.~\eqref{eq:crawl_symmetry_penalty}
& $-0.75$ \\
Torque saturation
& $\sum_{i\in\mathcal{J}_{\mathrm{all}}}|\tau_i-\tau_i^{\mathrm{PD}}|$
& $-0.1$ \\
\bottomrule
\end{tabularx}
\end{table}
\vspace{-10pt}
\subsection*{(B)~Transition-Policy Reward Terms}

Table~\ref{tab:transition_rewards_appendix} lists the additional terms applied on top of the BeyondMimic objective. Define the filtered contact indicator $\chi_l(t)=\mathds{1}\!\left(
\max_{k\in\{0,\ldots,4\}}\|\mathbf{f}_{t-k,l}\|_2>1.0
\right),$
% \begin{equation}
% \chi_l(t)=\mathds{1}\!\left(
% \max_{k\in\{0,\ldots,4\}}\|\mathbf{f}_{t-k,l}\|_2>1.0
% \right),
% \label{eq:filtered_contact_indicator}
% \end{equation}
where $\mathbf{f}_{t,l}$ is the contact force on link $l$. We use $\mathcal{L}$ for all links except the ankles, wrists, and knees, and $\mathcal{L}_{\mathrm{nonfeet}}$ for all non-foot links.

\begin{table}[!htbp]
\centering
\caption{Additional reward terms for bidirectional motion tracking.}
\label{tab:transition_rewards_appendix}
\footnotesize
\setlength{\tabcolsep}{2.5pt}
\renewcommand{\arraystretch}{1.12}
\begin{tabularx}{\columnwidth}{@{}p{0.28\columnwidth}Xc@{}}
\toprule
\textbf{Term} & \textbf{Definition} & \textbf{Weight} \\
\midrule
Head-height tracking
& $\mathds{1}(110\leq t_c\leq149)\exp[-(h^{\mathrm{head}}-0.5)^2/0.3^2]$
& $0.5$ \\
Knee-height tracking
& $\mathds{1}(110\leq t_c\leq149)\exp[-25(h^{\mathrm{knees}}-0.25)^2]$
& $1.0$ \\
Foot-level shaping
& $\mathds{1}(90\leq t_c\leq149)\|\mathbf{g}_{xy}^{\mathrm{feet}}\|$
& $-0.4$ \\
Hand-level shaping
& $\mathds{1}(0\leq t_c\leq95)
(\sqrt{1-g_y^{\mathrm{RH}}}+\sqrt{1+g_y^{\mathrm{LH}}})$
& $-0.6$ \\
Undesired contacts
& $\sum_{l\in\mathcal{L}}\chi_l(t)$
& $-0.3$ \\
Squat-phase contacts
& $\mathds{1}(90\leq t_c\leq149)
\sum_{l\in\mathcal{L}_{\mathrm{nonfeet}}}\chi_l(t)$
& $-0.5$ \\
\bottomrule
\end{tabularx}
\end{table}

% We used the following PPO parameters listed in Table \ref{tab:PPO_value_appendix} to train the policies for variable-height walking, crawling, and bidirectional transitions.
% \vspace{-10pt}
% \begin{table}[H]
% \centering
% \caption{PPO hyperparameters used for walking, crawling, and tracking training.}
% \label{tab:PPO_value_appendix}
% \footnotesize
% \renewcommand{\arraystretch}{0.95}

% \begin{tabular*}{\columnwidth}
% {@{\extracolsep{\fill}}lr@{\hspace{1.1cm}}}
% \toprule
% \textbf{Parameter} & \textbf{Value} \\
% \midrule
% Value loss coefficient         & $1.0$ \\
% Value loss clip                & $0.2$ \\
% Entropy coefficient            & $0.005$ \\
% Epochs per iteration           & $5$ \\
% Mini-batches per iteration     & $4$ \\
% Simulation steps per iteration & $20$ \\
% Decimation                     & $4$ \\
% Parallel environments          & $8192$ \\
% Learning rate                  & Adaptive \\
% Discount factor $\gamma$       & $0.99$ \\
% GAE parameter $\lambda$        & $0.95$ \\
% Target KL divergence           & $0.01$ \\
% \bottomrule
% \end{tabular*}
% \end{table}

% Let the bibliography continue immediately after the appendix (no page break).
\bibliographystyle{IEEEtran}
\bibliography{refs}

\end{document}